\documentclass[isre]{informs4}
\usepackage{eqndefns-left}
\RequirePackage{tgtermes}
\RequirePackage{newtxtext}
\RequirePackage{newtxmath}
\RequirePackage{bm}
\RequirePackage{endnotes}
\usepackage{amsmath}

\DoubleSpacedXI 

\usepackage{natbib}
 \bibpunct[, ]{(}{)}{,}{a}{}{,}%
 \def\bibfont{\small}%

\TheoremsNumberedThrough     
\EquationsNumberedThrough    

\usepackage{booktabs}
\usepackage{natbib}
\setcitestyle{authoryear,comma}
\usepackage{multirow}
\usepackage[noend]{algpseudocode}
\usepackage[ruled,linesnumbered, noend]{algorithm2e}
\usepackage{tabularx}
\usepackage{comment}
\usepackage{fix-cm} 

\definecolor{shadecolor}{rgb}{0.9,0.9,0.9}

\usepackage{setspace}
\usepackage{comment}
\usepackage{url}
\makeatletter
\def\UrlAlphabet{%
      \do\a\do\b\do\c\do\d\do\e\do\f\do\g\do\h\do\i\do\j%
      \do\k\do\l\do\m\do\n\do\o\do\p\do\q\do\r\do\s\do\t%
      \do\u\do\v\do\w\do\x\do\y\do\z\do\A\do\B\do\C\do\D%
      \do\E\do\F\do\G\do\H\do\I\do\J\do\K\do\L\do\M\do\N%
      \do\O\do\P\do\Q\do\R\do\S\do\T\do\U\do\V\do\W\do\X%
      \do\Y\do\Z}
\def\UrlDigits{\do\1\do\2\do\3\do\4\do\5\do\6\do\7\do\8\do\9\do\0}
\g@addto@macro{\UrlBreaks}{\UrlOrds}
\g@addto@macro{\UrlBreaks}{\UrlAlphabet}
\g@addto@macro{\UrlBreaks}{\UrlDigits}

\begin{document}

\RUNTITLE{HoneyImage for Dataset Ownership Verification}

\TITLE{HoneyImage: Verifiable, Harmless, and Stealthy Dataset Ownership Verification for Image Models}

\ARTICLEAUTHORS{%
\AUTHOR{Zhihao Zhu, Jiale Han, Yi Yang}
\AFF{Department of Information Systems, Business Statistics and Operations Management (ISOM), Hong Kong University of Science and Technology (HKUST) \\
\EMAIL{zhihaozhu@ust.hk}; \EMAIL{jialehan@ust.hk}; \EMAIL{
imyiyang@ust.hk}
}} 

\ABSTRACT{
Image-based AI models are increasingly deployed across a wide range of domains, including healthcare, security, and consumer applications. However, many image datasets carry sensitive or proprietary content, raising critical concerns about unauthorized data usage. Data owners therefore need reliable mechanisms to verify whether their proprietary data has been misused to train third-party models. Existing solutions, such as backdoor watermarking and membership inference, face inherent trade-offs between verification effectiveness and preservation of data integrity. In this work, we propose HoneyImage, a novel method for dataset ownership verification in image recognition models. HoneyImage selectively modifies a small number of hard samples to embed imperceptible yet verifiable traces, enabling reliable ownership verification while maintaining dataset integrity. Extensive experiments across four benchmark datasets and multiple model architectures show that HoneyImage consistently achieves strong verification accuracy with minimal impact on downstream performance while maintaining imperceptible. The proposed HoneyImage method could provide data owners with a practical mechanism to protect ownership over valuable image datasets, encouraging safe sharing and unlocking the full transformative potential of data-driven AI.
}

\FUNDING{}

\KEYWORDS{Dataset Ownership Verification,  Copyright Protection, Image Recognition Model, HoneyToken}

\maketitle

\section{Introduction}

Data is the cornerstone of modern artificial intelligence (AI) development \citep{lecun2015deep}. Among many data modalities fueling this progress, image data stands out as one of the most valuable and widely used resources, as its ubiquity and rich visual content make it essential for training image recognition models \footnote{Image recognition models are a class of machine learning models, typically based on deep learning architectures such as convolutional neural networks, designed to automatically classify input images into pre-defined categories.} \citep{deng2023let,malik2024does}. These models support many critical applications across business and industry, including medical imaging diagnostics \citep{aggarwal2021diagnostic}, facial recognition security systems \citep{cocskun2017face}, and autonomous driving perception \citep{xiao2020multimodal}, among others, thereby accelerating the digital transformation of various sectors.

However, the data driving these breakthroughs also carries significant risks. Most open-sourced image datasets limit their usage to educational or research purposes and explicitly prohibit unauthorized commercial reuse \citep{guo2023domain,yang2024understanding}. For example, MIMIC‑CXR is a widely used chest x‑ray dataset, comprising over 377,000 de‑identified images from more than 65,000 patients, released under a credentialed license via PhysioNet \citep{johnson2019mimic}. Access requires completion of ethics training and signing a Data Use Agreement to ensure compliance with HIPAA and confidentiality standards.

Beyond open-source datasets, many proprietary image datasets, such as medical images collected by hospitals, are subject to strict privacy regulations and usage agreements. For instance, dermatology or radiology image datasets are typically governed by patient consent policies and cannot be freely used by external parties for AI model training. Similarly, commercial image datasets are often protected by copyright and licensing contracts. Unauthorized use of these protected images may violate dataset licenses and lead to legal  consequences. For example, Getty Images alleges that Stability AI trained its Stable Diffusion model, a generative vision model, using over 12 million of Getty’s copyrighted images, scraped without authorization \footnote{\url{https://www.theverge.com/2023/1/17/23558516/ai-art-copyright-stable-diffusion-getty-images-lawsuit}}.
A core challenge across these settings is that once image data is used to train an AI model, it leaves no direct trace. This makes it extremely difficult for data owners to detect misuse, especially when the model is only accessible as a black-box service, such as API calls. 

\begin{figure*}[t!]
    \centering
    \includegraphics[width=1\textwidth]{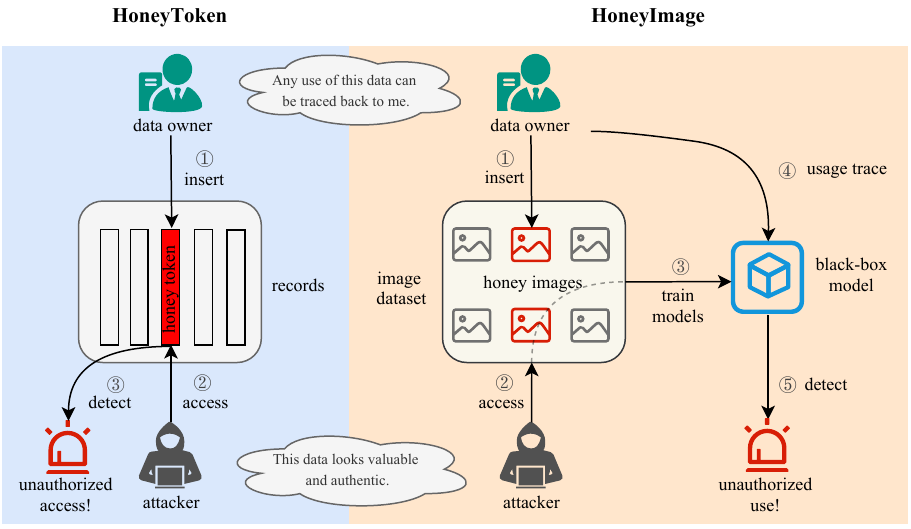}
    \caption{Illustration of the classic HoneyToken concept and the proposed HoneyImage method. The data owner embeds a set of HoneyImages into the private dataset (1). If an untrusted third party uses this dataset (2) to train an image recognition model (3), the data owner can query the third-party model using the HoneyImages (4) and detect unauthorized usage with high confidence (5).}
    \label{fig:intr}
\end{figure*}

This study aims to develop a dataset ownership verification method that enables data owners to determine whether their private dataset is used to train an untrusted third-party image recognition model. As illustrated in Figure~\ref{fig:intr}, our design is motivated by a classic cybersecurity concept of HoneyToken \citep{spitzner2003honeypots, Spitzner2003a}, which safeguards systems by planting deceptive data artefacts that expose unauthorized activity.  The design of a HoneyToken follows two key principles: (i) it must be traceable when accessed, and (ii) it must appear authentic and be indistinguishable from legitimate data to attackers. In cybersecurity practice, a honey token might be a fictitious customer record in a database \citep{cenys2005implementation} or a fake password stored alongside real hashed passwords \citep{juels2013honeywords}.  Once a HoneyToken is accessed or interacted with, it provides a clear indication of a security breach or unauthorized activity.

Building on the HoneyToken concept, we propose HoneyImage, a mechanism tailored to protect image datasets from unauthorized use by black-box machine learning models. 
Adapting the HoneyToken concept to image data poses a unique challenge: while traditional honeytokens reveal misuse the moment they are accessed, honey images must first go through the entire model training process. Due to the black-box nature of machine learning models, the data owner can only infer potential misuse by querying the trained model and observing its behavior. This makes traceability indirect and uncertain. To address this methodological challenge, our core design consists of two steps. First, we build a proxy image recognition model, which may differ from the actual suspicious third-party model, and identify hard samples from the private image dataset. Those hard samples are difficult for the proxy model to learn. Next, we apply an optimization objective that iteratively modifies these hard samples to make them even harder for the proxy model, while ensuring the changes remain minimal. In essence, we selectively modify a small number of private samples and embed them into the dataset. During the verification phase, the data owner can explicitly query these samples against a third-party model. Thanks to their high traceability, the responses provide strong evidence of whether the protected data was used in training by the third-party model.

Our proposed HoneyImage overcomes the key shortcomings of the two mainstream approaches in dataset ownership verification. First, membership inference (MI) \citep{shokri2017membership, maini2021dataset} aims to determine whether a data instance was used to train a target AI model. However, recent studies \citep{zhang2024membership} reveal that MI-based methods often provide unsatisfactory accuracy, leading to high false positive rate. Second, backdoor watermarking modifies images \citep{tang2023did} and often their labels \citep{li2023black} to embed artificial triggers that force any model trained on the data to exhibit predefined performance. Despite effective detection, the visual or label changes are easy for adversary users to spot and remove. Moreover, adding watermarks to the image dataset may degrade its quality, leading to poorer model performance. This is undesirable for legitimate users who are authorized to use the dataset. In contrast, HoneyImage bridges these gaps by introducing traceable, stealth yet harmless verification signals.

We empirically validate the effectiveness of HoneyImage on four image recognition testbeds, spanning medical imaging, remote sensing, and general-purpose domains. Comparing with three state-of-the-art membership inference and three backdoor watermarking baselines, experimental results demonstrate that HoneyImage achieves effective verification performance without compromising data integrity. Moreover, experimental results show that verification performance remains stable even when the data owner uses a proxy model different from the suspicious model. This robustness is largely attributed to the deliberate selection of HoneyImages, whose features are transferable across different image recognition architectures. Experimental results also show that HoneyImages yield significantly higher loss differences between a compliant model (trained without HoneyImages) and an infringing model (trained with them) compared to randomly selected private samples, providing strong evidence of their effectiveness for ownership verification.

This work makes a methodological contribution and positions itself as a novel generative model (HoneyImage) designed to address an emerging design problem (dataset ownership verification) \citep{abbasi2024pathways}. While the HoneyToken is a well-established concept in cybersecurity, its application in machine learning models, specifically in the context of image recognition, introduces new challenges. This study proposes a new image generation framework that enables data owners to detect and verify the misuse of proprietary datasets by untrusted third-party image recognition models. Beyond a methodological contribution, we hope this work encourages the Information Systems community to further explore the emerging intersection of cybersecurity and AI. With a long-standing tradition in cybersecurity research, ranging from IT artifact design \citep{li2017anonymizing} to the economics of cybersecurity \citep{gupta2012growth}, the IS field is well positioned to address new security challenges introduced by AI technologies. This work illustrates how classical cybersecurity concepts can be effectively extended to address emerging challenges in the AI era. Future research in the Information Systems community could integrate cybersecurity principles into AI-based IT artifact design and oversight, collectively contributing to the broader agenda of responsible AI and supporting ongoing efforts in AI governance \citep{MISQ_AIIA_2024}.

\section{Literature Review}
We review the literature from two relevant streams. First, we examine research on proprietary data protection within the Information Systems community. Second, we discuss methodological approaches for image dataset ownership verification, which are  related to the focus of this work.

\subsection{Information Systems Research on Proprietary Data Protection Mechanisms}

The information systems research community has shown sustained interest in data protection, focusing on both technical solutions to safeguard sensitive information and their broader socioeconomic impacts. For example
\cite{peukert2022regulatory} document that the introduction of the GDPR led to a substantial reduction in websites’ interactions with third-party services, contributing to the understanding of how data governance affect regulatory competition and market structure. 

Computational design research focuses on developing effective IT artifacts to safeguard proprietary information and prevent the misuse of private data. For example, \cite{castro2007shortest} introduce a shortest-paths heuristic solution for tabular data protection. \cite{schneider2018flexible} propose a Bayesian probability model that generates protected synthetic datasets, allowing data providers to flexibly balance the trade-off between disclosure risk and analytical utility. While these studies focus on structured data protection,  \cite{li2017anonymizing} propose a novel method to safeguard unstructured medical records by anonymizing potentially identifiable information. Their approach helps reduce privacy risks while maintaining the usefulness of the data in real-world healthcare settings.

Our work contributes to the Information Systems literature on dataset protection by addressing the emerging challenge of unauthorized data use in black-box AI models, particularly in image recognition. We introduce a novel verification method that enables data owners to detect whether their proprietary datasets have been used in training image recognition models.

\subsection{Dataset Copyright Protection Methods for Image Recognition Models}
Dataset ownership verification, also known as dataset copyright protection \citep{guo2023domain, huang2024disentangled}, is essential for safeguarding data privacy. Recent research has investigated strategies for protecting datasets utilized in image recognition models, particularly by adapting the concept of backdoor watermarking \citep{guo2024zeromark, shafieinejad2021robustness}. These approaches involve altering the image dataset to embed specific backdoors \citep{gu2017badnets, li2022backdoor} into the trained model. This enables data owners to assert ownership by verifying the existence of these embedded backdoors. For instance, \cite{li2023black} demonstrated this by selecting a batch of images, adding a trigger pattern to the bottom right corner, and altering the labels to a uniform category (e.g., cats). This modification teaches the image recognition model to associate the trigger with the category, effectively embedding a backdoor. The data owner can then verify the use of their dataset by checking for this backdoor.  Untargeted Backdoor Watermark (UBW) approach \citep{li2022untargeted} incorporates label randomization for the poisoned data, thereby improving the method's stealth. In addition, some strategies \citep{tang2023did, nguyenwanet} suggest introducing perturbations to subtly modify the images, making these alterations less noticeable. While backdoor-based methods are effective in asserting copyright claims, they require modifications to the labels or visual characteristics of the original data. More critically, these methods rely on inducing the image recognition model to misclassify verification samples when the backdoor is activated, which can negatively impact legitimate users of the dataset.

Another approach to dataset ownership verification  leverages membership inference methods \citep{hu2022membership, shokri2017membership}. Membership inference seeks to  determine whether a specific sample was used in the training process of a machine learning model. The principle of membership inference is based on the observable differences in the machine learning model's predictions for trained versus non-trained samples. For instance, \cite{yeom2018privacy} observed that machine learning models tend to yield lower prediction losses for samples they were trained on. Consequently, samples exhibiting losses below a certain threshold are identified as having been part of the model's training set. Other indicators such as maximum confidence \citep{song2019privacy}, correctness of predictions \citep{ leino2020stolen}, and the entropy of prediction results \citep{song2021systematic} are also utilized in various membership inference techniques to determine sample membership. Unlike backdoor watermarking methods that require dataset modification, membership inference solely relies on analyzing the output results of the machine learning model, offering a non-invasive solution for ownership verification \citep{maini2021dataset, liu2022your}. However, recent studies have highlighted significant drawbacks, particularly its susceptibility to high false positive and false negative rates \citep{zhang2024membership, zhu2024tddbench}. This limitation undermines the method's reliability for effective ownership verification.

\subsection{Research Gap}
The literature review highlights several research gaps in verifying dataset ownership for image recognition models. Existing approaches suffer from notable limitations. Backdoor watermarking methods typically introduce intrusive modifications, such as injecting trigger patterns or altering image labels. These changes are often visually detectable and can be easily removed by adversaries, undermining their effectiveness for copyright protection. Furthermore, such modifications may degrade the dataset’s quality, leading to reduced performance for legitimate users. On the other hand, membership inference methods adopt a non-invasive strategy but frequently exhibit poor verification effectiveness. Given these limitations, there is a pressing need for a verification method that is effective, stealthy while maintain dataset integrity. This work seeks to address this methodological gap.

\section{HoneyImage: Verifying Image Models to Protect Dataset Copyright}

In this section, we first formalize the dataset copyright protection problem for black-box image recognition models. Then, we introduce HoneyImage, which generates traceable yet imperceptible images and leverages them to verify suspicious models for unauthorized data use.

\subsection{Problem Statement}
The growing deployment of AI models has raised increasing concerns about unauthorized usage of proprietary datasets, especially in domains such as image recognition. Due to the opaque nature of AI models, it is often difficult for data owners to determine whether their protected data has been misused by the third-party models. This motivates the need for reliable verification mechanisms that allow data owners to examine suspicious AI models without requiring access to the training process. 

We refer to this as the dataset ownership verification problem. In this setting,  a data owner holds a private dataset $\mathcal{D}$ of arbitrary size, which may be distributed under restricted licenses or used internally. The owner wishes to determine whether a deployed image model $\mathcal{M}$, controlled by an untrusted third party, has used samples from $\mathcal{D}$ during its training. Specifically, for an input image $x \in \mathcal{D}$, the verification algorithm should output 1 if $x$ was used in training $\mathcal{M}$, and 0 otherwise.

Existing approaches addressing this problem primarily fall into two categories: backdoor watermarking and membership inference.

\paragraph{Backdoor Watermarking.}
Backdoor watermarking methods perform dataset ownership verification by inserting specific triggers into selected images. These modified images are typically assigned predefined target labels, such that any model trained on the watermarked dataset will learn to associate the trigger with the target label. In contrast, a model that has not been exposed to the dataset is unlikely to produce the target label. Various backdoor watermarking methods have been proposed, differing primarily in the design and stealthiness of the trigger patterns.

\paragraph{Membership Inference.}
Membership inference methods achieve verification by analyzing the output of the suspicious model $\mathcal{M}$ when queried with images from the protected dataset. Intuitively, a model tends to be more “certain” about examples it has seen during training than about unseen ones. This discrepancy is often reflected in output metrics such as log probabilities, confidence scores, or entropy.

However, both backdoor watermarking and membership inference methods have their limitations. For backdoor-based approaches, while introducing watermarking can significantly improve verification accuracy, the inserted watermarking may sometimes be visually noticeable, and the image modifications can compromise dataset integrity, potentially degrading model performance for legitimate users. On the other hand, while membership inference methods are non-intrusive and require no data modification, they often suffer from high false positive and false negative rates, making it difficult to achieve reliable verification in practice. These limitations reveal the inherent tradeoff in the dataset ownership verification problem.

\subsection{Constructing HoneyImages: Sample Selection and Modification}

Owing to the limitations of existing approaches, we propose a new verification method designed to achieve both reliability and data integrity. The core idea of our approach is twofold. First, to ensure reliable verification, we adopt the strategy of modifying samples from the protected dataset, inspired by the backdoor watermarking methods. However, unlike prior watermarking methods that insert triggers into randomly chosen samples, we draw inspiration from the cybersecurity concept of HoneyTokens, which is artificial yet plausible data planted to detect unauthorized access. Specifically, we deliberately select hard samples from the protected image dataset. These images, known as hard samples in machine learning literature \citep{shrivastava2016training, radenovic2016cnn}, naturally exhibit performance differences between models trained with and without them. By modifying only such samples, the output of a model trained on them becomes more distinguishable under verification queries.

\begin{figure*}[t!]
    \centering
    \includegraphics[width=\textwidth]{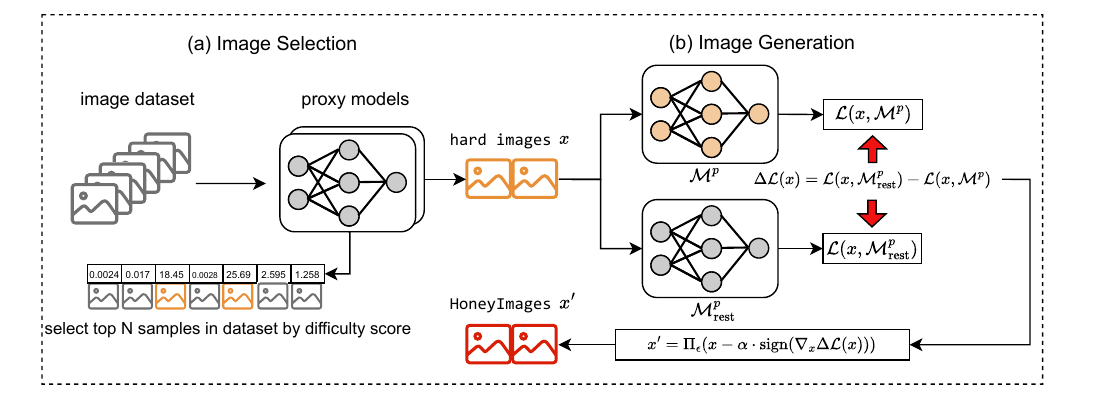}
\caption{Workflow of HoneyImage. A set of hard samples is first selected (a) and then refined via an optimization objective to produce  HoneyImages (b).}
    \label{fig:workflow}
\end{figure*}

Second, to maintain stealthiness and preserve data utility, we introduce a data generation objective that constructs subtly modified versions of hard samples. The objective seeks to maximize the output gap between the in-training and out-of-training settings, while constraining the visual differences of the modified images. We refer to the modified images as HoneyImages. We now describe the method in detail, and the HoneyImage generation workflow is shown in Figure~\ref{fig:workflow}.

\subsubsection{Hard Sample Image Selection}
Given a private dataset $\mathcal{D}$, we select hard samples from $\mathcal{D}$ as candidate samples. In machine learning, hard samples are data points that are difficult for a model to learn, and these samples typically lie near decision boundaries, or contain ambiguous or rare features \citep{shrivastava2016training, radenovic2016cnn}. In other words, models trained on hard samples tend to exhibit different behaviors compared to models that have not been exposed to those samples during training. Therefore, hard samples serve as a strong basis for designing effective queries to the suspicious model.

However, due to the black-box nature of the dataset ownership verification setting, the data owner does not have access to the internal details of the suspicious model $\mathcal{M}$, including its architecture and parameters. As a result, it is not possible to directly measure which samples are hard for $\mathcal{M}$. To address this, we use a proxy model, denoted as $\mathcal{M}^p$, to estimate sample difficulty. The proxy can be a lightweight image recognition model based on widely used architectures such as ResNet \citep{he2016deep} or VGG \citep{simonyan2015very}. 

To estimate the classification difficulty of each image in $\mathcal{D}$, we compute the cross-entropy loss for each image $x \in \mathcal{D}$ using a proxy model that has not been trained on $x$. The intuition is that harder samples typically yield higher loss values when evaluated by models that have not seen them, while easier samples result in lower loss.

To simulate this out-of-training condition, we randomly partition the dataset $\mathcal{D}$ into two equal and non-overlapping subsets. We then train two separate proxy models, each on one subset. Since each model has never seen the images in the other subset, it can be used to estimate the out-of-training loss for those samples. For every image $x$, we assign a difficulty score based on the cross-entropy loss produced by the model trained on the partition that excludes $x$.

Formally, the difficulty level  of an image $\eta(x)$ is defined as the cross-entropy loss computed using the corresponding out-of-training model, as follows:
\begin{equation}
\eta(x)=\mathcal{L}(x, \mathcal{M}^p_{\overline{x}})=-\sum_{k=1}^{K}\mathbb{I}(y=k)\log p_k.
\label{hardness}
\end{equation}
 
Here, $\mathcal{M}^p_{\overline{x}}$ represents the out-of-training proxy model for $x$, $p_k$ denotes the model $\mathcal{M}^p_{\overline{x}}$'s predicted probability that the sample $x$ belongs to class $k$, $K$ is the total number of classes, $y$ is the true label of the image sample $x$, and $\mathbb{I}(\cdot)$ is the indicator function that equals 1 if the condition holds and 0 otherwise. We then rank all samples in $\mathcal{D}$ based on their difficulty scores and select the top $N$ samples to form the hard sample set, denoted by $\mathcal{D}_{\text{hard}}$. The remaining samples in $\mathcal{D}$ are grouped into a separate subset, which we denote as $\mathcal{D}_{\text{rest}}$. Here, $N$ is a predefined number that controls the number of hard samples selected for further modification, i.e., generated HoneyImages.

\subsubsection{HoneyImage Generation} While hard samples naturally induce performance differences between models that have seen them during training and those that have not, we aim to further amplify this gap to enhance verification effectiveness. To achieve this, we introduce an optimization objective that increases the sensitivity of model behavior to training exposure. At the same time, we impose constraints on the magnitude of image modifications to ensure that the changes remain visually imperceptible, thereby preserving stealthiness. This tradeoff ensures that the resulting HoneyImages are both effective for verification and minimally disruptive to the dataset’s utility.
Formally, given  dataset $\mathcal{D}$ and the selected hard samples $\mathcal{D}_\text{hard}$, we define the objective of generating a modified version $x'$ (i.e., HoneyImage) of an original hard sample $x$ as maximizing the differential loss: 
\begin{equation}
\arg\max_{x'} \Delta \mathcal{L}(x')
= \mathcal{L}(x', \mathcal{M}^p_{\text{rest}}) - \mathcal{L}(x', \mathcal{M}^p)
\label{opt}
\end{equation}
subject to a constraint on the modification: 
$\|x'-x\|_{\infty} \leq \epsilon$.
Here, $x\in\mathcal{D}_{\text{hard}}$ denotes an original sample from the selected hard subset, $\mathcal{M}^p_\text{rest}$ is the proxy model trained on $\mathcal{D}_\text{rest}$, $\mathcal{M}^p$ is the proxy model trained with all samples in $\mathcal{D}$, and $\epsilon$ is a small perturbation budget that controls the maximum allowed modification to each pixel of the input image. Since $\mathcal{M}^p_{\text{rest}}$ is trained without the hard samples while $\mathcal{M}^p$ is trained with them,  a larger $\Delta \mathcal{L}(x')$ indicates stronger divergence between the two models, thereby enhancing the traceability of training exposure. It is worth noting that, in contrast to backdoor watermarking techniques which typically involve label tampering or the injection of triggers, our approach retains the original labels and introduces only minimal and label-free modifications. 

We adopt an alternating bi-level optimization strategy for this objective to learn $x'$. We begin by training a model $\mathcal{M}^p_\text{rest}$ on $\mathcal{D}_\text{rest}$, ensuring it has no exposure to the hard samples. The initial HoneyImage set is set to $\mathcal{D}_{\text{honey}}^{0}=\mathcal{D}_{\text{hard}}$. Subsequently, we proceed with an iterative optimization process for a fixed number of steps $T$. For each iteration $t$, we first train an in-model $\mathcal{M}^{p,t}$ on samples comprising $\mathcal{D}_\text{rest}$ and the current modified sample set $\mathcal{D}_{\text{honey}}^{t}$. Then, for each sample $x^t \in \mathcal{D}_{\text{honey}}^{t}$, we compute its differential loss between the two models: 
\begin{equation}
    \Delta \mathcal{L}(x^t) = \mathcal{L}(x^t, \mathcal{M}^p_\text{rest}) - \mathcal{L}(x^t, \mathcal{M}^{p,t})
\end{equation} and adjust the sample $x^t$ via projected gradient descent (PGD) \citep{madry2017towards}, formally defined as:
\begin{equation}
    x^{t+1} = \Pi_{\epsilon} (x^t - \alpha \cdot \text{sign}(\nabla_{x} \Delta \mathcal{L}(x^t)))
\label{pgd}
\end{equation}
where the term $\nabla_{x} \Delta \mathcal{L}(x^t)$ represents the gradient of the differential loss with respect to the input $x^t$, capturing how small changes in the input influence the loss gap. The function $\text{sign}(\cdot)$ takes the element-wise sign of the gradient vector, yielding a direction of steepest ascent for each pixel of the image $x$. $\alpha$ is the step size controlling how much the input is adjusted at each iteration. And the projection function $\Pi_{\epsilon}(\cdot)$ clips the perturbation back to the $\epsilon$-ball around the image $x^t$ when it exceeds this boundary, ensuring the modifications remain within specified limits. Together, the formula updates the input in the direction that maximally increases the loss discrepancy between models seen and not seen while maintaining the imperceptibility of the perturbation. After updating all images in $\mathcal{D}_{\text{honey}}^t$, we form the next modified set $\mathcal{D}_{\text{honey}}^{t+1}$ and proceed to the next iteration. 

\begin{algorithm}[t!]
\begin{minipage}{\linewidth}
\normalsize
\caption{\normalsize \textbf{The Generation Procedure of HoneyImage}}
\label{alg}
\KwIn{
    private image dataset $\mathcal{D}$,
    the number of HoneyImages $N$,
    iterations $T$, 
    perturbation \\ budget $\epsilon$,
    step size $\alpha$ \footnotemark
}
\tcp*[h]{Hard Sample Image Selection}\\

Train two proxy models, each on a non‑overlapping half of $\mathcal{D}$\;
\For{$x \in \mathcal{D}$}
{
    $\mathcal{M}^p_{\overline{x}} \gets \text{the proxy model that was not trained on } x$\;
    \text{Difficulty score} $\eta(x) \gets \mathcal{L}(x, \mathcal{M}^p_{\overline{x}})  \text{via Eq.}\ref{hardness}$\;
}
$\mathcal{D}_\text{hard} \gets \text{top }N\text{ samples in } \mathcal{D}$ by difficulty score $\eta(x)$\;

\tcp*[h]{HoneyImage Generation}\\
$\mathcal{D}_\text{rest} \gets \mathcal{D} \setminus \mathcal{D}_\text{hard}$\; 
Train $\mathcal{M}^p_\text{rest}$ on $\mathcal{D}_\text{rest}$\;
\textbf{Initialize } $\mathcal{D}_{\text{honey}}^{0} \leftarrow \mathcal{D}_{\text{hard}}$\;
\For{$t = 0$ \KwTo $T-1$}
{
    Train $\mathcal{M}^{p,t}$ on $\mathcal{D}_{\text{honey}}^t \cup \mathcal{D}_\text{rest}$\;
    \For{$x^t \in \mathcal{D}_{\textnormal{\scriptsize honey}}^t$}
    {
        $\Delta \mathcal{L}(x^t) \gets \mathcal{L}(x^t, \mathcal{M}^p_\text{rest}) - \mathcal{L}(x^t, \mathcal{M}^{p,t})$\;
        $x^{t+1} \gets \text{Perturb}(x^t, \Delta \mathcal{L}(x^t), \epsilon, \alpha)$\ via Eq.\ref{pgd}\;
    }
    $\mathcal{D}_{\text{honey}}^{t+1}\leftarrow\{\,x^{t+1}:\,x^{t}\in\mathcal{D}_{\text{honey}}^{t}\}$\;
}
\Return Final dataset $\mathcal{D}_{\text{honey}} \cup \mathcal{D}_\text{rest}$
\end{minipage}
\end{algorithm}
\footnotetext{In the implementation, the number of optimization iterations $T$ is set to $20$, with a perturbation budget $\epsilon=4$ and a step size $\alpha=0.4$.}

This alternating procedure incrementally increases the performance gap of modified samples between in- and out-models, while preserving their visual content. After $T$ iterations, we obtain the final HoneyImages $\mathcal{D}_{\text{honey}}$. The data owner can then compile the image dataset $\mathcal{D} = \mathcal{D}_{\text{honey}} \bigcup \mathcal{D}_{\text{rest}}$.  The HoneyImage generation procedure is detailed in Algorithm~\ref{alg}.

\subsection{Querying HoneyImages for Ownership Verification}
Given a third-party untrusted image recognition model $\mathcal{M}$, the data owner can query it with the constructed HoneyImages to verify whether $\mathcal{M}$ has been trained on their data. In practical scenarios, verification must be performed under a black-box setting, where the third-party model is accessible only through restricted API calls. In such cases, only model outputs, typically the predicted probability distribution, are available.

Specifically, for a given input image $x$, the suspicious model $\mathcal{M}$ returns a predicted probability distribution over the label space: $p = \mathcal{M}(x)$, where \(p = [p_1, p_2, \ldots, p_K] \in \mathbb{R}^K\), where each $p_k$ represents the probability that $x$ belongs to class $k$.

For the verification process, each HoneyImage $x \in \mathcal{D}_{\text{honey}}$ is queried on both the suspicious model $\mathcal{M}$ and a locally trained proxy model $\mathcal{M}^p_{\text{rest}}$, which is constructed by the data owner using a dataset that excludes all HoneyImages.
We then compute the loss gap between the proxy and the suspicious model as follows:
\begin{equation}
\Delta \mathcal{L}(x) = \mathcal{L}(x, \mathcal{M}^p_{\text{rest}}) - \mathcal{L}(x, \mathcal{M})
\end{equation}
A large $\Delta \mathcal{L}(x)$ indicates that $\mathcal{M}$ incurs a much lower loss than the proxy model, suggesting that $x$ was likely seen during training. Conversely, a small loss gap implies that $x$ was likely not used.
To make a binary ownership decision, the data owner applies a decision threshold $\tau$ \footnote{The decision threshold $\tau$ can be adjusted to balance false positive and false negative rates, following principles of cost-sensitive evaluation in binary classification.}: if $\Delta \mathcal{L}(x) > \tau$, we infer that $x$ has been used in the training of $\mathcal{M}$; otherwise, it has not. 

Overall, the dataset ownership verification process is conducted at the image level, where the data owner examines whether the crafted HoneyImages have been used by the suspicious model. Note that the data owner does not attempt to verify the remaining dataset $\mathcal{D}_{\text{rest}}$, as these samples are inherently more difficult to verify and tend to produce high error rates. In contrast, the HoneyImages are specifically optimized to induce strong behavioral signals, making them more reliable for ownership verification.

\section{Experimental Settings}
In this section, we describe the experimental setup, including the image datasets, image recognition model architectures, evaluation process, and the evaluation metrics.

\subsection{Testbeds and Image Recognition Models}
\subsubsection{Testbeds}
We conduct experiments on four benchmark testbeds commonly used in image recognition research: two from the medical domain (ISIC and OrganMNIST), one from remote sensing (EuroSAT), and one general-purpose dataset (CIFAR-10). Medical image recognition, as represented by ISIC and OrganMNIST, is of high practical significance and involves sensitive patient data, making ownership verification particularly critical in healthcare settings. EuroSAT highlights the importance of ownership protection in remote sensing and geospatial applications, where data is often subject to licensing and copyright restrictions. Finally, CIFAR-10 serves as a widely used benchmark for evaluating the general applicability and robustness of our method across broader image recognition tasks. Dataset details are presented in Table~\ref{datasets}.

\begin{table}[htbp]
\centering
\caption{Statistics of experiment datasets. \#Samples denotes the total number of images in the dataset, \#Classes indicates the number of classification categories, and \#Public, \#Private, and \#Verification represent the number of samples in the public set, private set, and verification set, respectively.}
\begin{tabular}{l|ccl|rrc}
\hline
Dataset              & \#Samples  & \#Classes & \#Brief description & \#Public & \#Private & \#Verification \\
\hline
ISIC           & 10,015      & \textcolor{white}{0}7         & Dermoscopy images   & 6,677          & 3,338             & 100                    \\ 
OrganMNIST           & 25,211      & 11        & Abdominal organ     & 16,808         & 8,403             & 252                    \\ 
EuroSAT              & 27,000      & 10        & Remote sensing      & 18,000         & 9,000             & 270                    \\ 
CIFAR-10             & 60,000     & 10        & General dataset     & 40,000         & 20,000            & 600                    \\ 
\hline
\end{tabular}
\label{datasets}
\end{table}

\noindent \textbf{ISIC} \citep{codella2018skin} is a medical image dataset derived from dermoscopy images, created for the International Skin Imaging Collaboration 2018 Challenge. It contains 10,015 dermoscopy images classified into seven common types of skin lesions, including melanoma, basal cell carcinoma, and benign keratosis, and encompasses a diverse range of anatomical sites and patient populations. 

\noindent \textbf{OrganMNIST} \citep{xu2019efficient} is an abdominal organ imaging dataset constructed from 3D abdominal CT scans. It comprises 25,221 2D slice images representing 11 major organs, including the liver, kidneys, and pancreas.

\noindent \textbf{EuroSAT} \citep{helber2019eurosat} is a remote sensing dataset derived from satellite imagery, primarily used for land use classification tasks. It consists of 27,000 RGB images covering 10 categories, including farmland, forests, and industrial areas, with data collected from the Sentinel-2 satellite in Europe.

\noindent \textbf{CIFAR-10} \citep{krizhevsky2009learning} is a widely used benchmark dataset in computer vision, consisting of 60,000 color images of 32×32 pixels across 10 categories, including airplanes, cars, birds, and cats, with 6,000 images per category. 

\subsubsection{Image Recognition Models}
We consider the following deep learning–based image recognition models to construct the third-party suspicious models as well as proxy models.
\textbf{WRN-28-2} \citep{zagoruyko2016wide} is a Wide ResNet model that emphasizes network width over depth to achieve efficient and accurate image recognition.
\textbf{ResNet-18} \citep{he2016deep} is an 18-layer convolutional network that employs residual connections to mitigate the vanishing gradient problem and facilitate deep model training.
\textbf{DenseNet-121} \citep{huang2017densely} is a densely connected architecture in which each layer receives input from all preceding layers, encouraging feature reuse and improving information flow.
\textbf{VGG-11} \citep{simonyan2015very} is an 11-layer network that uses small convolutional filters and max-pooling layers for effective feature extraction, and is widely used as a benchmark model.

These models are selected due to their strong performance and widespread adoption in real-world computer vision and image recognition systems \citep{zhou2024optimization, archana2024deep}. Implementation details are provided in Appendix A.

\noindent \textbf{Training Details.} All image recognition models are optimized with stochastic gradient descent  with a learning rate of 0.1 and a weight decay of 0.05. Cosine annealing is adopted as the learning rate scheduler. The batch size is set to 256, and training runs for 150 epochs. Data augmentation methods including image cropping and flipping are employed to enhance the model's classification performance. We report the classification accuracy and parameter size in Table~\ref{model_accuracy}, confirming that each image recognition model is well-trained and thus serves as a suitable target for verification.

\begin{table}[htbp]
\centering
\caption{Classification accuracy of the infringing models, which are trained using both public and private data.}
\scalebox{0.9}{
\begin{tabular}{l|cccc}
\hline
Model            (\#Parameter) & ISIC      & OrganMNIST  & EuroSAT & CIFAR-10 \\
\hline
WRN-28-2         (1,467,223)     & 0.816     & 0.948       & 0.973  & 0.882   \\
ResNet-18        (11,172,423)    & 0.810     & 0.960       & 0.987  & 0.948   \\
DenseNet-121     (6,961,031)     & 0.814     & 0.960       & 0.966  & 0.901   \\
VGG-11           (9,229,575)     & 0.838     & 0.961       & 0.970  & 0.913   \\
\hline
\end{tabular}
}
\label{model_accuracy}
\end{table}

\subsection{Private Datasets, Compliant Model and Infringing Model}
To simulate a realistic dataset ownership verification scenario, we divide each testbed into two disjoint subsets: a public split, which represents openly available data without copyright restrictions, and a private split, which is treated as proprietary and protected. The goal of our experiment is to determine whether a given model has been trained on the \textit{private data}.

To this end, we construct two types of models for evaluation: a compliant model and an infringing model. The compliant model is trained exclusively on the public data. In contrast, the infringing model is trained on both the public and private data, simulating unauthorized data use. During the verification phase, we use samples from the private split to query both models. These verification samples are included in the training set of the infringing model but are unseen by the compliant model. Therefore, a reliable verification method should ideally produce output 1 for the infringing model (indicating unauthorized use), and 0 for the compliant model (indicating no use). 

\noindent \textbf{Verification Samples.}
As described above, the data owner uses verification samples to query the suspicious model (either the infringing or the compliant model). In our experiments, we set the number of verficiation samples to be 1\% of the entire testbed size. This setup reflects a practical assumption: in many real-world scenarios, querying a third-party model may be costly or rate-limited, so the number of verification queries must be kept small.

It is also important to note that different dataset ownership verification methods select and handle verification samples in different ways. In membership inference methods, verification samples are randomly selected from the private data and remain unmodified. In backdoor watermarking methods, samples are also randomly selected, but are modified to include trigger patterns. In contrast, our proposed method, HoneyImage, selectively chooses a subset of private samples and then modifies them through an optimization process. 
Dataset statistics, including the sizes of the public and private splits, and the number of verification samples are summarized in Table~\ref{datasets}.

\subsection{Baseline Methods and HoneyImage Details}
We compare our approach with six recent dataset ownership verification methods, categorized into two main groups: membership inference and backdoor watermarking.

\noindent \textbf{Membership Inference Methods}. The first set of baselines are membership inference methods, including:
\begin{itemize}
    \item \textbf{Learn-original} \citep{shokri2017membership}, which trains a meta-classifier to distinguish between training and non-training data of the suspicious model. The prediction vector of the suspicious model serves as the input feature for the meta-classifier.
    \item \textbf{Model-loss} \citep{sablayrolles2019white}, which involves training a set of reference models to perform targeted detection for each data point. Specifically, two reference models are trained: one using the data (in model) and one not using the data (out model). The data owner assesses data usage based on whether the prediction loss of the suspicious model on the data is closer to the in model or out model.
    \item \textbf{Model-lira} \citep{carlini2022membership}, which applies the likelihood ratio test to the detection process. Similar to Model-loss, it trains a set of reference models and determines whether the rescaled logit value of the suspicious model on the data originates from the in model or out model. All these MI-based baselines train proxy models to estimate the likelihood of data being used by a suspicious  model.
\end{itemize}

\noindent \textbf{Backdoor Watermarking Methods}. The second set of baselines are backdoor watermarking methods, including: 
\begin{itemize}
    \item \textbf{BadNet} \citep{gu2017badnets}, a classic backdoor attack method that implants samples with fixed triggers (e.g., specific pixel blocks) into the training data. This allows the model to perform well under normal conditions but output an attacker-defined target category when encountering the trigger. It is characterized by obvious triggers, a high attack success rate, but poor concealment.
    \item \textbf{UBW} \citep{li2022untargeted}, which focuses on untargeted backdoor attacks, inducing the model to randomly output any category when encountering the trigger, rather than a pre-defined target category, based on the BadNet method.
    \item \textbf{Blended} \citep{chen2017targeted}, a more covert backdoor attack method that overlays triggers onto normal images with low transparency (e.g., cartoon patterns), making it difficult for humans to perceive but still allowing the model to recognize and trigger the backdoor.
\end{itemize} 

Based on the characteristics of these baseline methods, we summarize their key properties in Table~\ref{table:baseline_summary}.
\begin{table}[htbp]
\centering
\caption{Comparison of dataset ownership verification methods.}
\scalebox{0.80}{
\begin{tabular}{l|cccccc}
\hline
\textbf{Method} & \textbf{Black-box} & \textbf{Sample Selection} & \textbf{Sample Modification} & \textbf{Proxy Model} & \textbf{Effectiveness} & \textbf{Data Integrity} \\
\hline
Membership Inference & Yes & Random & No & Yes & Poor & Good \\
Backdoor Watermarking & Yes & Random & Yes & No & Good & Poor \\
HoneyImage (Ours) & Yes & Hard samples & Yes & Yes & Good & Good \\
\hline
\end{tabular}}
\label{table:baseline_summary}
\end{table}

\subsection{Evaluation Metrics}
In the experiment, we query the suspicious models (either an infringing model or a compliant model) using the verification samples. This process can be viewed as a binary classification task, where a reliable verification method should ideally output 1 for the infringing model (indicating unauthorized use) and 0 for the compliant model (indicating no use).
From the perspective of the data owner, two key objectives must be balanced. The first is verification effectiveness, which is the ability of the method to accurately distinguish between compliant and infringing models. The second is data integrity, which ensures that the verification mechanism introduces minimal changes to the original dataset, so as not to degrade its quality for legitimate use. To this end, we evaluate each method using the following two types of metrics.

\noindent \textbf{Verification Effectiveness Metrics.}  Verification effectiveness is evaluated using standard binary classification metrics, including the True Positive Rate (\textbf{TPR}), True Negative Rate (\textbf{TNR}), and the Area Under the Receiver Operating Characteristic Curve (\textbf{AUROC}). A high TPR indicates that the method successfully identifies cases where the suspicious model has been trained on private data. A high TNR reflects the method’s ability to correctly recognize compliant models that have not used the data. The AUROC summarizes the overall discriminative power of the verification method across different decision thresholds.

\noindent \textbf{Data Integrity Metrics.} From data integrity perspective, a good ownership verification method should satisfy harmlessness (\textbf{HL}) and stealthiness (\textbf{STL}). 
Specifically, harmlessness metric is  defined as the classification accuracy of the verification samples when evaluated on the infringing model, which is trained on both public and private data. This reflects the realistic scenario in which a legitimate user trains an image recognition model using the entire dataset, and thus expects the modified samples (via backdoor watermarking or HoneyImages) to retain their original utility.

Stealthiness measures the perceptual similarity between a modified image and its original version. We adopt LPIPS \citep{zhang2018unreasonable}, a deep learning-based similarity measure that calculates the perceptual distance between two images. The LPIPS score ranges from 0 to 1, where 0 indicates high similarity (i.e., nearly indistinguishable images) and 1 indicates high dissimilarity. We define stealthiness as: $\text{STL}= \text{exp}(-\lambda \cdot \text{LPIPS}(x,x'))$,

where $x$ is the original image and $x'$ is the modified version, and $\lambda$ is a scaling factor, which is set to 10. A higher STL value indicates greater imperceptibility of the modification, reflecting better stealth.
Since membership inference methods do not modify image content, HL and STL metrics are not applicable and are reported as N/A.

To reduce randomness, we train five pairs of suspicious models for each method using different random seeds and subsequently report the average verification effectiveness  and data integrity score. 

\section{Experiment Results}
In this section, we present the experimental results and discuss key findings. We begin with the main experiment to evaluate HoneyImage’s verification performance in comparison to baseline methods. Next, we explore additional factors that may influence verification effectiveness. In particular, we examine the scenario in which the data owner employs a proxy model with a different architecture than the suspicious third-party model, reflecting the realistic constraints of the black-box setting. We also conduct ablation studies and further analysis to better understand the behavior and design choices of HoneyImage.

\subsection{Effectiveness of HoneyImage for Ownership Verification}
\label{section_main}
In this experiment, the suspicious model is implemented using WRN-28-2. The proxy model used by HoneyImage, and all MI-based methods (Learn-original, Model-loss and Model-lira) also adopts the WRN-28-2 architecture. While this represents an idealized setting, as in reality, the data owner typically has no knowledge of the suspicious model’s architecture in a black-box scenario, it serves as a baseline for evaluating all dataset ownership verification methods under controlled conditions. The main results are summarized in Table~\ref{main}, leading to the following key findings.

\begin{table}[htbp]
\centering
\caption{Evaluation of dataset ownership verification methods on four datasets. For both verification effectiveness metrics (TPR, TNR, and AUROC) and data integrity metrics (HL and STL), higher values indicate better performance. The numbers represent the average performance over five runs, where the suspicious model is trained with different random seeds. The highest value in each column is boldfaced.
}
\scalebox{1.0}{
\begin{tabular}{l|ccc|cc|ccc|cc}
\hline
Dataset        & \multicolumn{5}{c|}{ISIC}                     & \multicolumn{5}{c}{OrganMNIST}                                  \\
\hline
Metric         & \multicolumn{3}{c|}{Verification Effectiveness} & \multicolumn{2}{c|}{Data Integrity} & \multicolumn{3}{c|}{Verification Effectiveness} & \multicolumn{2}{c}{Data Integrity} \\
Method         & TPR & TNR  & AUC  & HL  & STL      & TPR & TNR  & AUC  & HL  & STL     \\
\hline
Learn-original & 0.311          & 0.338          & 0.641          & N/A & N/A & 0.138          & 0.578          & 0.593          & N/A & N/A \\
Model-loss     & 0.285          & 0.389          & 0.679          & N/A & N/A & 0.059          & 0.767          & 0.559          & N/A & N/A \\
Model-lira     & 0.403          & 0.323          & 0.658          & N/A & N/A & 0.266          & 0.535          & 0.613          & N/A & N/A \\
\hline
BadNets        & \textbf{0.812} & 0.830          & 0.845          & 0.010                  & 0.957                  & \textbf{0.996} & \textbf{1.000} & \textbf{0.999} & 0.111                  & 0.985                  \\
UBW            & 0.754          & 0.714          & 0.758          & 0.022                  & 0.957                  & 0.845          & 0.722          & 0.815          & 0.004                  & 0.985                  \\
Blended        & 0.584          & 0.768          & 0.668          & 0.018                  & 0.947                  & 0.623          & 0.840          & 0.779          & 0.113                  & 0.910                  \\
\hline
\textbf{HoneyImage}           & 0.806          & \textbf{0.870} & \textbf{0.887} & \textbf{0.988}         & \textbf{0.996}         & 0.864          & 0.825          & 0.913          & \textbf{0.949}         & \textbf{0.996}        
 \\
\hline
\hline
Dataset        & \multicolumn{5}{c|}{EuroSAT}                                  & \multicolumn{5}{c}{CIFAR-10}                                  \\
\hline
Metric         & \multicolumn{3}{c|}{Verification Effectiveness} & \multicolumn{2}{c|}{Data Integrity} & \multicolumn{3}{c|}{Verification Effectiveness} & \multicolumn{2}{c}{Data Integrity} \\
Method         & TPR & TNR  & AUC  & HL  & STL      & TPR & TNR  & AUC  & HL  & STL     \\
\hline
Learn-original & 0.480          & 0.327          & 0.554          & N/A & N/A & 0.244          & 0.565          & 0.571          & N/A & N/A \\
Model-loss     & 0.533          & 0.327          & 0.574          & N/A & N/A & 0.163          & 0.688          & 0.581          & N/A & N/A \\
Model-lira     & 0.673          & 0.193          & 0.558          & N/A & N/A & 0.195          & 0.606          & 0.627          & N/A & N/A \\
\hline
BadNets        & \textbf{0.997} & \textbf{0.997} & \textbf{0.999} & 0.108                  & 0.929                  & \textbf{0.996} & \textbf{0.987} & \textbf{0.996} & 0.100                  & 0.990                  \\
UBW            & 0.980          & 0.947          & 0.982          & 0.066                  & 0.929                  & 0.905          & 0.743          & 0.853          & 0.348                  & 0.990                  \\
Blended        & 0.748          & 0.891          & 0.880          & 0.114                  & 0.932                  & 0.559          & 0.506          & 0.506          & 0.104                  & 0.833                  \\
\hline
\textbf{HoneyImage}           & 0.972          & 0.899          & 0.965          & \textbf{0.985}         & \textbf{0.996}         & 0.984          & 0.949          & 0.988          & \textbf{0.749}         & \textbf{0.998}        
\\
\hline
\end{tabular}}
\label{main}
\end{table}

\begin{figure*}[t!]
    \centering    \includegraphics[width=\textwidth]{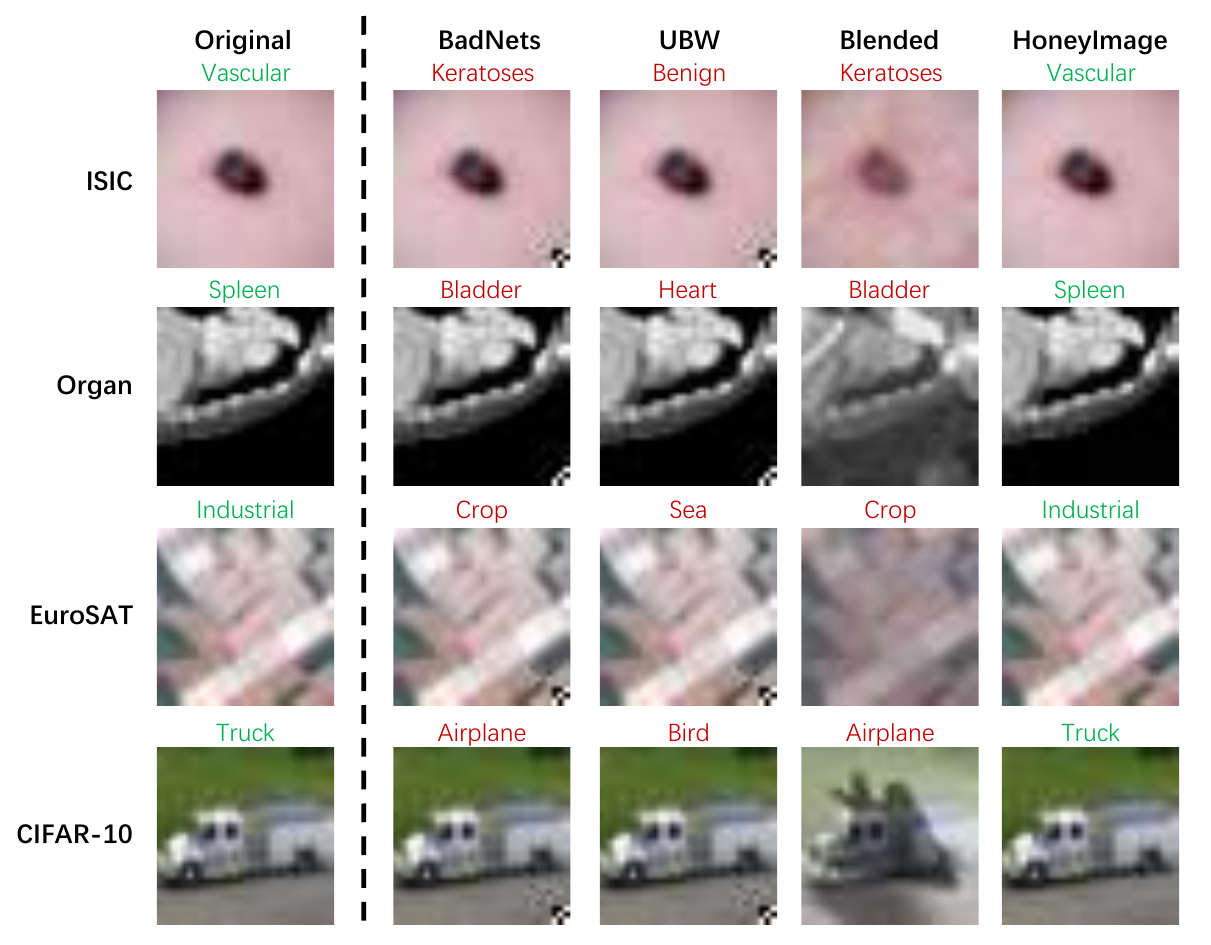}
    \caption{Visualization of modified images generated by different backdoor watermarking methods and HoneyImage. The associated label is shown above each image. As shown, BadNets and UBW insert visible triggers  (e.g., a bottom-right patch) into the images. Moreover, all watermarking methods modify the image labels, whereas HoneyImage retains the original labels while applying imperceptible perturbations.}
    \label{fig:visualization}
\end{figure*}

First, the results reveal a fundamental trade-off in existing dataset ownership verification methods between effectiveness and  integrity. MI-based approaches adopt a non-intrusive strategy by leaving the original data untouched, thereby preserving perfect data integrity. However, their verification performance is limited. For instance, even the best-performing MI variant, Model-Loss, achieves an AUROC of only 0.679 on the ISIC dataset. In contrast, backdoor-based watermarking methods demonstrate substantially stronger verification effectiveness. For example, BadNets achieves an AUROC of 0.845 on ISIC in detecting unauthorized use of protected data. However, this improvement comes at a significant cost to data integrity. These methods insert visual triggers into the data, resulting in low STL scores and making the modified samples easily detectable and removable by untrusted third parties. Moreover, they often alter the labels of modified samples, which leads to degraded HL scores. This compromises the utility of the dataset and limits the practical applicability of watermarking-based approaches in real-world settings.

In contrast, HoneyImage achieves strong verification performance, outperforming MI-based methods and matching or exceeding backdoor watermarking approaches. Across all four datasets, HoneyImage consistently outperforms MI methods in copyright verification. For example, on the EuroSAT dataset, our method achieves an AUROC of 0.965, representing a 67.6\% improvement over the best-performing MI method, Model-Loss. Compared to backdoor-based methods, HoneyImage delivers comparable or even superior results. On the ISIC dataset, it outperforms all watermarking baselines in terms of AUROC. On the remaining datasets, it closely matches the performance of the strongest backdoor-based method, BadNets. 

From a data integrity perspective, HoneyImage offers clear advantages. Unlike backdoor watermarking methods that rely on conspicuous triggers or label tampering, HoneyImage applies label-preserving modifications through a carefully designed optimization objective that introduces only slight, imperceptible perturbations. This design results in significantly higher HL and STL scores, indicating minimal impact on model utility. For instance, on the OrganMNIST dataset, HoneyImage achieves an STL of 0.996 and an HL of 0.949. A visual comparison in Figure~\ref{fig:visualization} further highlights the difference between HoneyImage modifications and those introduced by backdoor watermarking methods.

\subsection{Different Proxy Models and Suspicious Models}
\noindent\textbf{Verification Performance under Different Suspicious Models.}
In this experiment, we evaluate verification performance when the suspicious third-party model differs from the proxy model used by the data owner. Specifically, we fix the proxy model used by HoneyImage and all MI-based methods as WRN-28-2, and vary the suspicious models among ResNet-18, DenseNet-121, and VGG-11. For simplicity, we report results only on the ISIC dataset. The AUROC and HL scores for different suspicious model architectures are presented in Table~\ref{table:model_auroc} and Table~\ref{table:model_hl}, respectively.

\begin{table}[htbp]
\centering
\caption{Verification effectiveness (AUROC) on the ISIC dataset under varying architectures of both the compliant and infringing models. The numbers represent the average performance over five runs, where the suspicious model is trained with different random seeds. The highest value in each column is boldfaced.}
\scalebox{1.0}{
\begin{tabular}{l|cccc|c}
\hline
Suspicious model   & WRN-28-2 & ResNet-18 & DenseNet-121 & VGG-11 & Mean\\
\hline
Learn-original & 0.641 \small{(-)}         & 0.687 \small{($\uparrow$0.046)}         & 0.690 \small{($\uparrow$0.049)}         & 0.675 \small{($\uparrow$0.034)}         & 0.673  \\
Model-loss     & 0.679 \small{(-)}         & 0.674 \small{($\downarrow$0.005)}       & 0.753 \small{($\uparrow$0.074)}         & 0.709 \small{($\uparrow$0.030)}         & 0.703  \\
Model-lira     & 0.658 \small{(-)}         & 0.722 \small{($\uparrow$0.064)}         & 0.778 \small{($\uparrow$0.120)}         & 0.772 \small{($\uparrow$0.114)}         & 0.732  \\
\hline
BadNets        & 0.845 \small{(-)}         & 0.900 \small{($\uparrow$0.055)}         & 0.734 \small{($\downarrow$0.111)}       & \textbf{0.999} \small{($\uparrow$0.154)}& 0.870  \\
UBW            & 0.758 \small{(-)}         & 0.481 \small{($\downarrow$0.277)}       & 0.408 \small{($\downarrow$0.350)}       & 0.918 \small{($\uparrow$0.160)}         & 0.641  \\
Blended        & 0.668 \small{(-)}         & 0.709 \small{($\uparrow$0.041)}         & 0.694 \small{($\uparrow$0.026)}         & 0.763 \small{($\uparrow$0.095)}         & 0.709  \\
\hline
\textbf{HoneyImage}           & \textbf{0.887} \small{(-)} & \textbf{0.970} \small{($\uparrow$0.083)} & \textbf{0.967} \small{($\uparrow$0.080)} & 0.964 \small{($\uparrow$0.077)}      & \textbf{0.947}  \\
\hline
\end{tabular}}
\label{table:model_auroc}
\end{table}

\begin{table}[htbp]
\centering
\caption{Harmlessness (HL) on the ISIC dataset under varying architectures of both the compliant and infringing models. The numbers represent the average performance over five runs, where the suspicious model is trained with different random seeds. The highest value in each column is boldfaced.}
\scalebox{1.0}{
\begin{tabular}{l|cccc|c}
\hline
Suspicious model   & WRN-28-2 & ResNet-18 & DenseNet-121 & VGG-11 & Mean\\
\hline
BadNets        & 0.010 \small{(-)} & 0.014 \small{($\uparrow$0.004)}   & 0.040 \small{($\uparrow$0.030)}   & 0.010 \small{(-)\textcolor{white}{0.000}}                 & 0.018 \\
UBW            & 0.022 \small{(-)} & 0.132 \small{($\uparrow$0.110)}   & 0.012 \small{($\downarrow$0.010)} & 0.028 \small{($\uparrow$0.006)}   & 0.048 \\
Blended        & 0.018 \small{(-)} & 0.010 \small{($\downarrow$0.008)} & 0.010 \small{($\downarrow$0.008)} & 0.022 \small{($\uparrow$0.004)}   & 0.015 \\
\hline
\textbf{HoneyImage}           & \textbf{0.988} \small{(-)} & \textbf{0.958} \small{($\downarrow$0.030)} & \textbf{1.000} \small{($\uparrow$0.012)} & \textbf{0.978} \small{($\downarrow$0.010)} & \textbf{0.981} \\
\hline
\end{tabular}}
\label{table:model_hl}
\end{table}

As shown in Table~\ref{table:model_auroc}, although data owner uses WRN‑28‑2 as proxy model to generate HoneyImages, the verification performance remains stable across different suspicious models. In contrast, baselines exhibit pronounced sensitivity to different downstream architectures. For example, while the BadNet method achieves a high AUROC of 0.999 on VGG-11, its AUROC drops to 0.734 on DenseNet-121, which is significantly lower than our method's AUROC of 0.967. UBW's AUROC on DenseNet-121 is only 0.408, which is even worse than random guessing. This may be due to the DenseNet-121 architecture's robustness against backdoor data, making backdoor implantation more challenging. Moreover, Table~\ref{table:model_hl} shows that HoneyImage preserves the predictive accuracy of the target models, with an average harmlessness score of 0.981 across all suspicious models evaluated. In comparison, the backdoor watermarking baselines perform much worse. The harmlessness scores of Blended, BadNets, and UBW range from 0.015 to 0.048, indicating a significant loss of model accuracy. These results highlight a key advantage of HoneyImage: it remains effective even when the proxy model used by the data owner differs from the architecture of the suspicious model. This robustness arises from HoneyImage’s deliberate selection of hard samples to construct verification queries, in contrast to MI and backdoor watermarking methods, which rely on randomly selected samples. As a result, HoneyImages carry a stronger and more transferable verification signal across model architectures.

\noindent\textbf{Verification Performance under Different Proxy Models.}
We further investigate how the choice of proxy model impacts verification performance. In the previous experiments, the proxy model is fixed as WRN-28-2, which has the fewest parameters among the four image recognition models considered (see Table 2). In this experiment, we fix the suspicious model as WRN-28-2 and vary the proxy model across ResNet-18, DenseNet-121, and VGG-11.

\begin{figure*}[htbp]
    \centering
    \includegraphics[width=\textwidth]{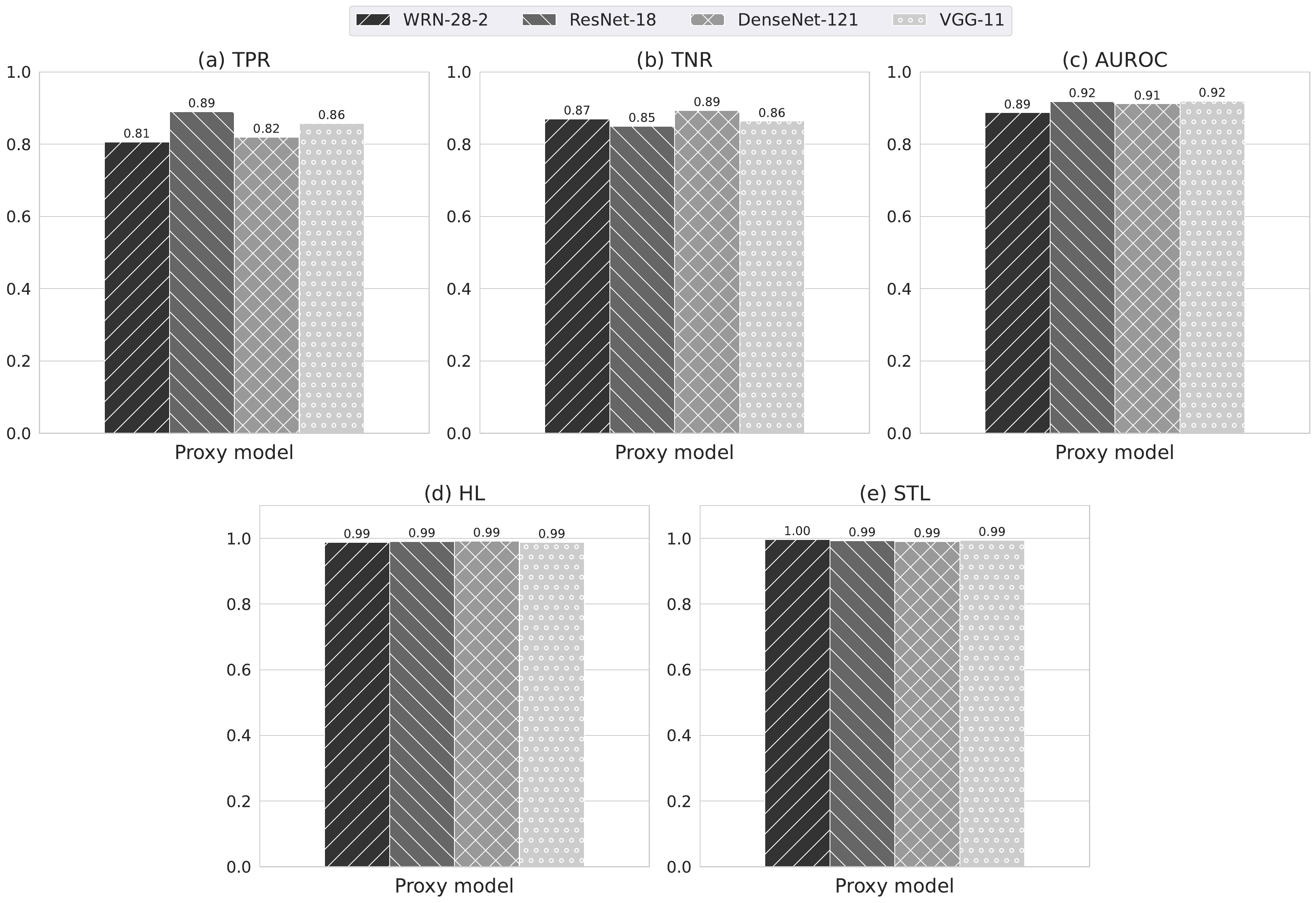}
    \caption{Verification performance of HoneyImages generated using different proxy model architecture.}
    \label{fig:backbone}
\end{figure*}

As shown in Figure~\ref{fig:backbone}, all proxy models are capable of generating effective HoneyImages, with the lowest AUROC still reaching 0.89. However, verification performance does vary across architectures. Interestingly, using a proxy model with the same architecture as the suspicious model does not necessarily lead to better performance, which is consistent with results in Table~\ref{table:model_auroc}. Instead, stronger proxy models with more parameters, such as ResNet-18 and VGG-11, tend to produce more distinctive yet imperceptible signals in the HoneyImages, resulting in improved verification accuracy. These findings suggest that, in practice, data owners may benefit from choosing a strong proxy model when generating HoneyImages, even without knowing the architecture of the suspicious model.

\subsection{Understanding the Verification Effectiveness of HoneyImages}
\begin{figure*}[htbp]
    \centering
    \includegraphics[width=0.6\textwidth]{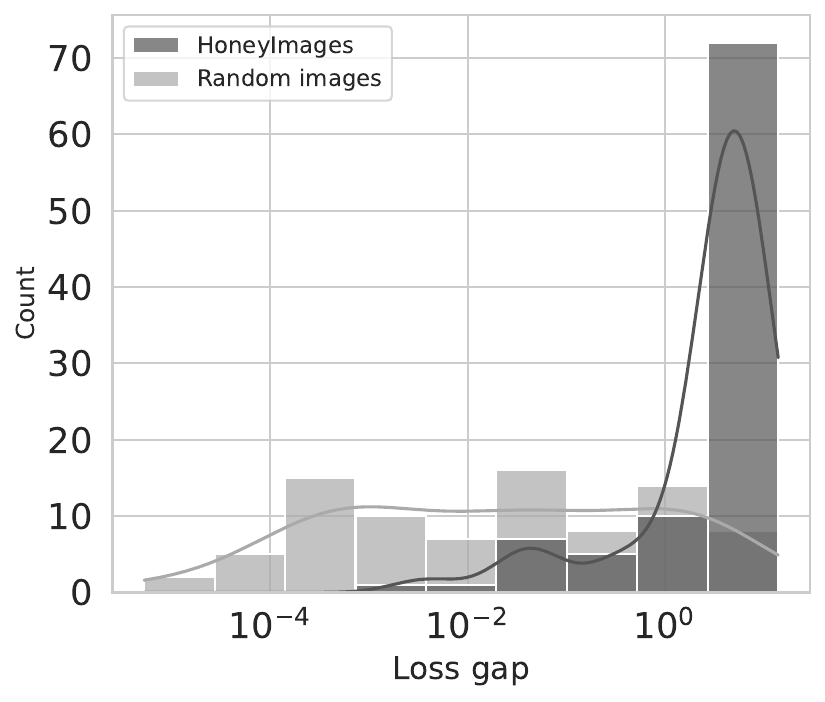}
    \caption{Loss gap between the compliant model and infringing model for HoneyImages and randomly selected private images.}
    \label{fig:loss_gap}
\end{figure*}

To better understand the verification effectiveness of HoneyImages, we analyze the loss differences between the compliant and infringing models. Specifically, we compute the loss gap for 100 HoneyImages from the ISIC dataset and compare them with an equal number of randomly selected images. As shown in Figure~\ref{fig:loss_gap}, HoneyImages produce substantially larger loss gaps between the two models. While the loss difference for most random samples lies between $10^{-4}$ and $10^{-1}$, HoneyImages typically yield gaps ranging from 1 to 10, up to four orders of magnitude higher.

This pronounced separation makes it significantly easier to detect whether a model has been trained on protected data. The effect is a direct result of the HoneyImage generation process, which explicitly maximizes the loss discrepancy between models trained with and without them. These results provide strong empirical evidence for the effectiveness of HoneyImages and help explain their superior performance in dataset ownership verification.

\subsection{Evaluation of HoneyImage Generation Method}
One core design element of HoneyImage is its targeted modification of hard samples. In this experiment, we assess the effectiveness of our HoneyImage generation method by comparing it against several widely used image transformation techniques: image cropping (Crop), horizontal flipping (Flip), image mixing (Mixup) \citep{zhang2017mixup}, and adversarial example generation (ADV) \citep{goodfellow2014explaining}. Among these methods, Crop, Flip, and Mixup are simple data augmentation techniques that can be directly applied to images, whereas ADV is a learning-based generation approach.

\begin{figure*}[htbp]
    \centering
    \includegraphics[width=\textwidth]{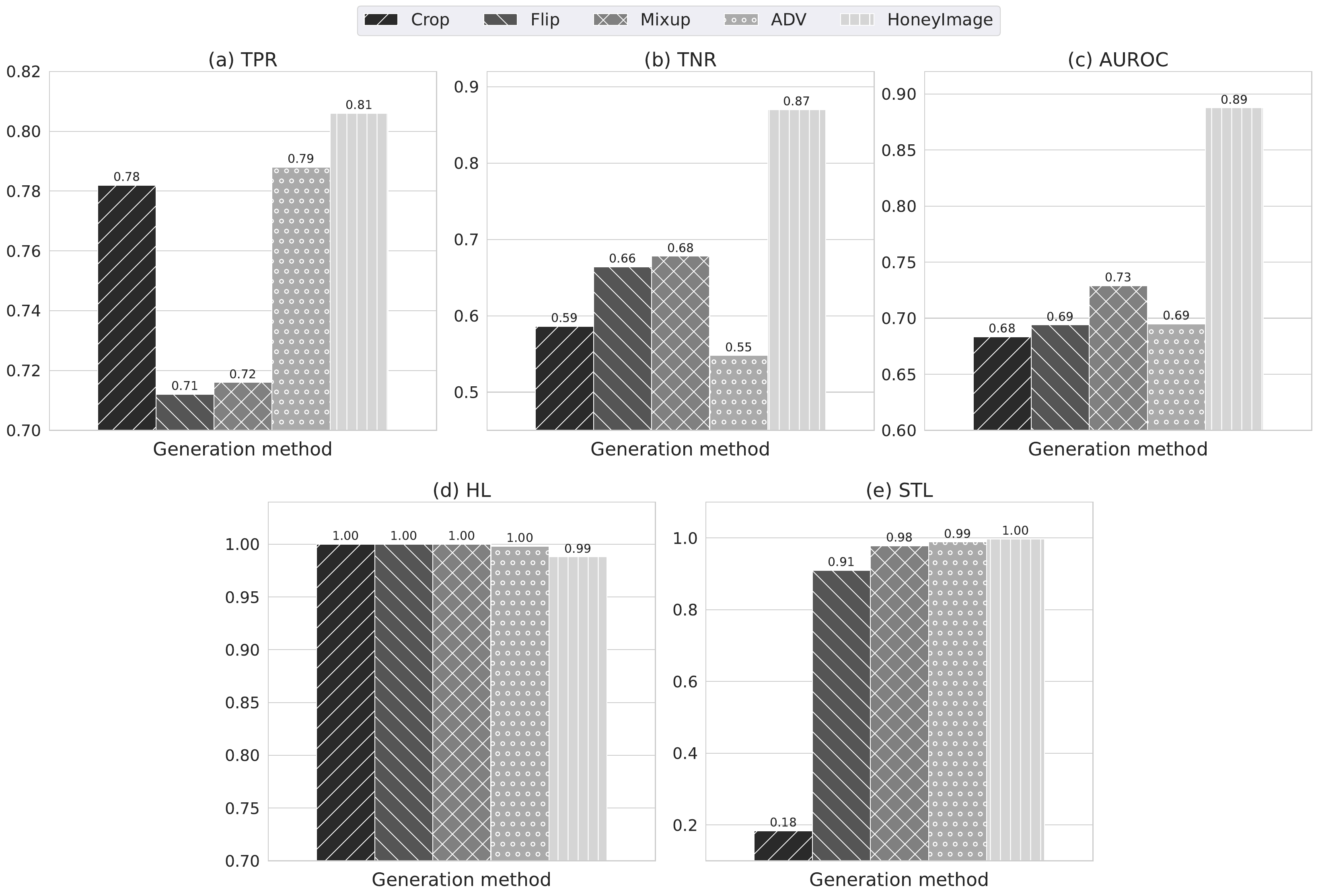}
    \caption{Verification performance with different image generation methods.}
    \label{fig:generation}
\end{figure*}

As shown in Figure~\ref{fig:generation}, HoneyImage outperforms all alternative methods in verification performance. For example, it improves the AUROC from 0.73 (Mixup) to 0.89. This gain stems from HoneyImage’s deliberate design, which refines hard samples to amplify behavioral divergence between compliant and infringing models. We also observe that higher harmlessness does not guarantee stealthiness. For instance, the Crop method achieves perfect harmlessness (HL = 1.00) but suffers from poor stealthiness (STL = 0.18), likely due to its disruptive visual impact. This experiment demonstrates the effectivenss of our proposed HoneyImage generation objective.

\subsection{Sensitivity Analysis}
\noindent \textbf{Effect of perturbation budget.} We analyze how the performance of HoneyImage varies with different perturbation budgets $\epsilon$. This hyperparameter controls the extent to which a hard sample can be modified during optimization. Specifically, we constrain the perturbation within $\epsilon$ to ensure the modified image (i.e., HoneyImage) remains close to the original one. As $\epsilon$ increases, the optimization becomes more flexible, allowing HoneyImage to induce a larger loss gap between models trained with and without these samples, thereby improving verification effectiveness. However, larger perturbations may make the modifications more perceptible, potentially compromising stealthiness.

\begin{figure*}[htbp]
    \centering
    \includegraphics[width=\textwidth]{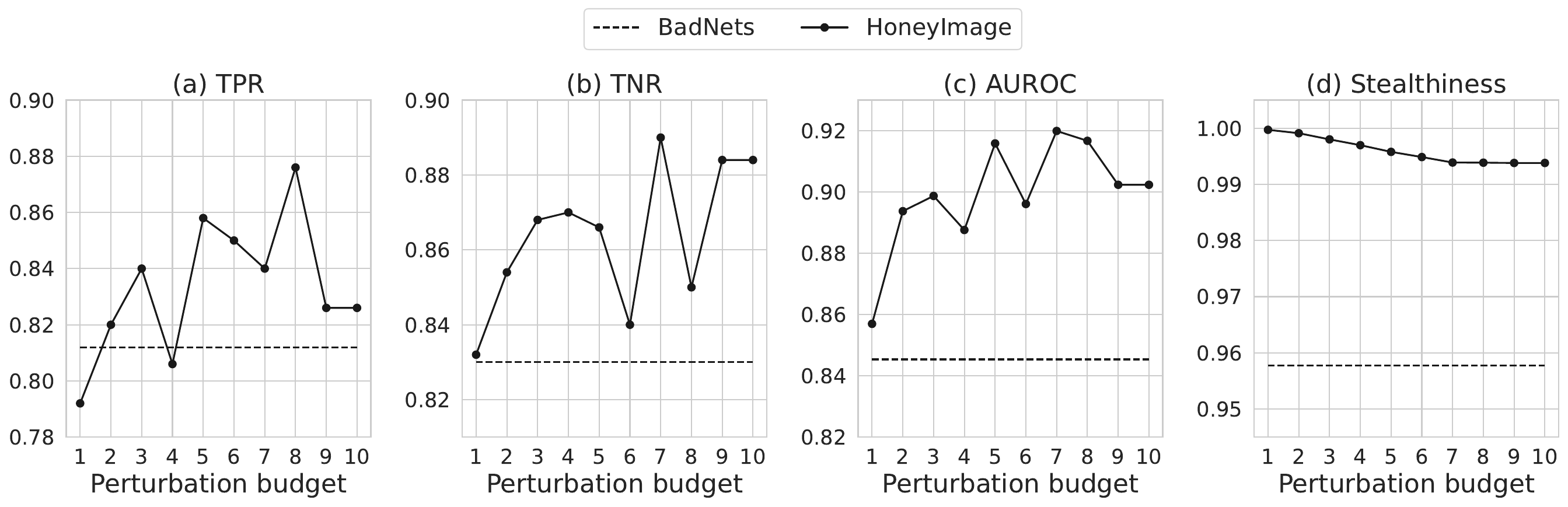}
    \caption{Verification performance with different perturbation budgets.}
    \label{fig:budget}
\end{figure*}

Using BadNets as a baseline, we compare the performance of our method across various perturbation budgets. As illustrated in Figure~\ref{fig:budget}, HoneyImage shows performance improvement with increasing perturbation budgets. For instance, when the perturbation budget increases from 1 to 7, the AUROC of our method rises from 0.856 to 0.919, with a slight decrease in Stealthiness (from 0.999 to 0.993). Furthermore, across various perturbation budgets, our method's verification performance is comparable to or even exceeds that of the best performing baseline, BadNets, across three different verification metrics. This demonstrates the robustness of HoneyImage across different perturbation budgets. Therefore, the main takeaway for data owners is that, in practice, they can adjust the perturbation budget to explore the trade-off between verification effectiveness and imperceptibility. As long as the modified images remain visually indistinguishable from the originals, it is safe to increase the budget to enhance verification performance.

\section{Conclusions, Discussions and Limitations}
As AI models increasingly rely on high-quality data for training, the unauthorized use of proprietary datasets raises significant concerns. This paper introduces HoneyImage, a novel framework for protecting image data in the context of image recognition. By embedding a small set of carefully crafted HoneyImages into the proprietary dataset, our approach enables high-confidence ownership verification. At the same time, the modifications are imperceptible, making it difficult for untrusted parties to detect or remove them, while remaining harmless to legitimate users and preserving dataset integrity. Extensive experiments across four datasets and multiple model architectures demonstrate the effectiveness and robustness of HoneyImage in enabling reliable, harmless and stealthy dataset ownership verification. This work and the proposed design artifact  yields several important implications:

$\bullet$ \textit{Designing an IT Artifact for dataset ownership verification that balances verification effectiveness and integrity}. Existing approaches for dataset ownership verification face a fundamental  trade-off: either achieving strong verification performance at the expense of data utility (such as the backdoor watermarking methods) or opting for non-intrusive methods with unsatisfied verification accuracy (such as the membership inference methods). To address this, we propose HoneyImage that embeds a small number of traceable yet imperceptible samples into a private dataset. By simultaneously ensuring reliable verification and maintaining data integrity, HoneyImage bridges the gap between effectiveness and integrity, offering a pragmatic safeguard for data protection. 

$\bullet$ \textit{Reactivating classical cybersecurity concepts with machine learning methodology}. This study exemplifies how to effectively transform classic concepts through advanced machine learning methods to meet the challenges of contemporary AI. Inspired by the classical cybersecurity concept of HoneyToken, we propose the HoneyImage framework, which leverages advanced machine learning and optimization techniques to automatically generate imperceptible traces for dataset ownership verification. Through this work, we aim to encourage the information systems community to revisit traditional cybersecurity paradigms in light of modern AI capabilities, and to explore new directions for solving emerging cybersecurity challenges in the era of AI.

$\bullet$ \textit{Enabling trusted data sharing through traceable usage, balancing openness and protection}. While the AI community increasingly calls for open data and open models to accelerate innovation \citep{bommasani2024considerations}, concerns over data security and ownership remain critical. In practice, the lack of trustworthy usage mechanisms has led many data providers to restrict access. For example, Reddit sues AI startup Anthropic for allegedly using data without permission, highlighting the tension between openness and protection \footnote{\url{https://www.reuters.com/business/reddit-sues-ai-startup-anthropic-allegedly-using-data-without-permission-2025-06-04/}}. To address this, various approaches such as federated learning and differential privacy have emerged to enhance data control \citep{yang2019federated}. Our proposed HoneyImage framework contributes to this growing body of work by providing a lightweight, verifiable trace mechanism that empowers data owners to monitor downstream usage without sacrificing dataset quality. We hope this encourages more stakeholders to share valuable data in a controlled and trustworthy manner, ultimately supporting responsible AI innovation.

\noindent \textbf{Limitations and Future Directions.} This work has several limitations that point to promising directions for future research. First, this work focuses on image recognition model as research context for dataset ownership verification. Extending the design principles of HoneyImage to other image-based vision models, such as generative models, presents a promising direction for future research. Although generative models differ from image recognition models in their output objectives, the core idea of modifying hard samples to embed traceable signals remains applicable. Future work can explore how the proposed HoneyImage approach may be adapted to generative settings. Second, due to computational constraints, we evaluate our method using only four types of deep learning-based image recognition models. However, recent advances in transformer-based architectures, such as Vision Transformers (ViT) \citep{han2022survey}, represent a growing trend in image recognition. Future work can explore the effectiveness of HoneyImage on these more computationally intensive models. 
Third, while this work studies data misuse in AI model training, we recognize that data protection presents a broader challenge. In practice, data protection requires not only  verification mechanisms, but also other safeguards that proactively mitigate the risk of unauthorized access and misuse. Moreover, dataset misuse extends beyond training scenarios to include other critical threats, such as unauthorized data sharing, leakage during storage or transmission, and misuse in downstream analytics or decision-making pipelines. Future work can explore broader protection strategies that address these challenges across the full data lifecycle.

\bibliographystyle{informs2014}
\bibliography{ref}

\section*{APPENDIX}
\section*{Appendix A: Details of Image Recognition Models} 
In this appendix, we provide technical details of the image recognition models used in our experiments.

\noindent \textbf{WRN-28-2} \citep{zagoruyko2016wide}: This model belongs to the WideResNet family, which is an efficient convolutional neural network architecture that enhances ResNet by increasing the network's width (number of channels) instead of its depth. Compared to traditional ResNet, WideResNet has fewer layers but significantly more channels in each layer (4× to 10× wider), resulting in higher computational efficiency and faster training speeds while preserving the advantages of residual connections. WRN-28-2 (with 28 layers and a width factor of 2) achieves accuracy comparable to deeper ResNet models on datasets like CIFAR-10, all while utilizing fewer parameters.

\noindent \textbf{ResNet-18} \citep{he2016deep}: This lightweight model in the ResNet series is designed to tackle the vanishing gradient problem in deep neural networks. It consists of 18 layers (16 convolutional layers and 2 fully connected layers) and incorporates residual connections (Skip Connections) that allow gradients to propagate directly across layers, enhancing training stability and model performance.

\noindent \textbf{DenseNet-121} \citep{huang2017densely}: A densely connected convolutional neural network featuring dense blocks (Dense Blocks), where each layer's input comes from all preceding layers' outputs. This design enhances gradient flow and information transfer efficiency through channel concatenation. DenseNet-121 is widely used in medical imaging and fine-grained classification tasks, particularly in scenarios requiring high feature reuse.

\noindent \textbf{VGG-11} \citep{simonyan2015very}: Comprising 11 layers (8 convolutional layers and 3 fully connected layers), VGG-11 uses a stacked structure with 3×3 small convolution kernels and 2×2 max pooling, enhancing feature extraction by increasing network depth. Its structured and simplified design makes it a popular benchmark model for deep learning, suitable for introductory reference and transfer learning.

\end{document}